\documentclass[a4paper,11pt]{article}

\usepackage{fourier}
\usepackage{color}
\usepackage{graphicx}
\usepackage{url}
\usepackage[affil-it]{authblk}
\usepackage{amsmath}
\usepackage{wrapfig}

\usepackage[T1]{fontenc}
\usepackage{times}

\let\OLDthebibliography\thebibliography
\renewcommand\thebibliography[1]{
  \OLDthebibliography{#1}
  \setlength{\parskip}{3pt}
}

\begin{document}

\title{Recurrent Super-Resolution Method for Enhancing Low Quality Thermal Facial Data}

\author[1]{David O'Callaghan\thanks{david.ocallaghan@xperi.com}}
\author[1]{Cian Ryan}
\author[1,2]{Waseem Shariff}
\author[1,2]{Muhammad Ali Farooq}
\author[1]{Joseph Lemley}
\author[2]{Peter Corcoran}
\affil[1]{Xperi Corporation, Galway, Ireland}
\affil[2]{School of Engineering, National University of Ireland, Galway}
\date{}
\maketitle
\thispagestyle{empty}

\begin{abstract}
The process of obtaining high-resolution images from single  or multiple low-resolution images of the same scene is of great interest for real-world image and signal processing applications. This study is about exploring the potential usage of deep learning based image super-resolution algorithms on thermal data for producing high quality thermal imaging results for in-cabin vehicular driver monitoring systems. In this work we have proposed and developed a novel multi-image super-resolution recurrent neural network to enhance the resolution and improve the quality of low-resolution thermal imaging data captured from uncooled thermal cameras. The end-to-end fully convolutional neural network is trained from scratch on newly acquired thermal data of 30 different subjects in indoor environmental conditions. The effectiveness of the thermally tuned super-resolution network is validated quantitatively as well as qualitatively on test data of 6 distinct subjects. The network was able to achieve a mean peak signal to noise ratio of 39.24 on the validation dataset for 4x super-resolution, outperforming bicubic interpolation both quantitatively and qualitatively. 
\end{abstract}

\section{Introduction}
Super-resolution (SR) is used to increase the details of a low-resolution (LR) image by applying statistical approaches, and various optimization techniques on either a sequence of images in a collective manner or a single image to generate high-resolution image. In this article, we have focused on developing an imaging pipeline to output super-enhanced thermal images for in-cabin driver monitoring systems. Primarily, with the advancements in safe and autonomous systems for vehicular technology, it is essential to monitor the driver’s activity for enabling enhanced security and safety features. This can be achieved by face localization, facial landmark detection, drowsiness and fatigue detection. In this work, particular focus is on developing a thermal super-resolution algorithm for enhancing the existing thermal imaging data captured from Video Graphics Array (VGA) uncooled prototype Long-wave Infrared (LWIR) camera developed under the Heliaus project \cite{heliaus}. Given recent developments in microbolometer technology, uncooled thermal imaging sensors are now less expensive and can be used for wide range real-world applications \cite{farooq2021object}. This sensor has added benefits as they can detect the thermal emissivity of objects and work independently of illumination conditions (i.e., a thermal camera can operate independently in any circumstance, including day and night), making them a more reliable source of data for vehicular in-cabin applications. The proposed SR algorithm integrates recurrent-based multi-image super-resolution (MISR) and single-image super-resolution (SISR), thus taking benefits from both the methods to produce optimal results. 

\section{Background}
\vspace{-0.2cm}
Image super-resolution is a popular research topic in the scientific community. And, given the challenges of thermal vision, super-resolution of thermal images is a difficult research problem. A number of useful thermal datasets have been released in recent years \cite{wang2010natural, espinosa2013new}. It is currently crucial to develop an effective neural network algorithm to super-resolve these images using available datasets. The deep back-projection network (DBPN) \cite{haris2018deep} was proposed for super-resolution of single visible images. Further improvements include up-sampling layers with a recurrent network \cite{haris2019recurrent}, an unsupervised approach using CycleGAN \cite{rivadeneira2020thermal}, Spatio-Temporal feature fusion deep neural network \cite{zhang2021infrared}, with different up-sampling and asymmetrical residual learning in the network \cite{patel2021thermisrnet}. Considering thermal image super-resolution is an open research topic, a CVPR workshop also included a thermal super-resolution challenge. The challenge was tackled by 9 separate teams, each with a unique solution \cite{rivadeneira2021thermal}. Recently a new approach was proposed to this problem that is based on a generative adversarial network with channel filtering mainly focused on super-resolution for power equipment \cite{haris2018deep}. This network outperformed the state-of-the-art in the reconstruction of thermal imaging images (of power equipment) by having a higher peak signal to noise ratio (PSNR) and structural similarity (SSIM) to the enhanced image.

\section{Methodology}
\vspace{-0.2cm}
\subsection{Data collection}
\vspace{-0.1cm}
The acquired thermal SR dataset was created after a series of data acquisitions which were conducted as part of the Heliaus project \cite{heliaus} in a driving simulator. The dataset consisted thermal images of 36 individuals (7 female and 29 male) using a VGA quality 640$\times$480 uncooled thermal camera based on microbolometer technology developed under the Heliaus project. The mean age of the individuals was 37.7 years and the standard deviation was 11.3 years. On average, 543 sequences of thermal images (each with a length of 10 frames) were extracted from the recording of each subject. This resulted in a dataset of 19,558 thermal image sequences. The data of 30 subjects were used for training and 6 subjects were used for testing. The effective train-test split ratio was 0.83:0.17.
\subsection{Neural Network Architecture}

\begin{wrapfigure}{r}{0.6\textwidth}

\vspace{-0.9cm}
\includegraphics[width=0.6\textwidth]{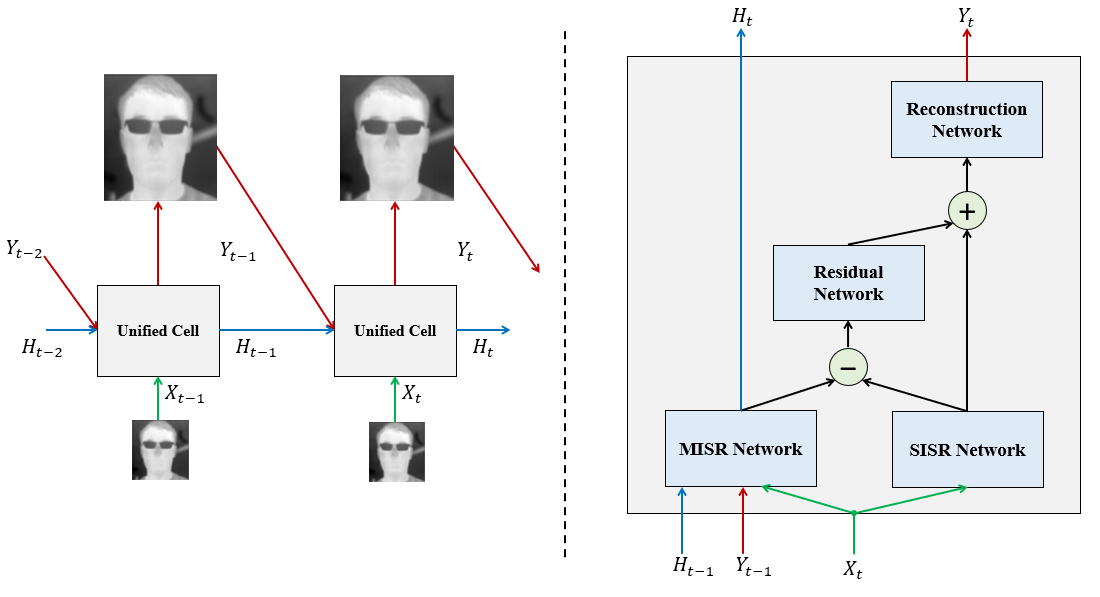}
\vspace{-0.7cm}
\caption{The unrolled recurrent network (left) and components of the unified cell (right) of our super-resolution network.}
\label{fig:network}
\vspace{-0.2cm}
\end{wrapfigure}

The proposed novel super-resolution architecture is illustrated in Figure \ref{fig:network}. In the left image, low-resolution images are combined with latent space feature maps computed at the previous time-step and the previous SR network output. Information is propagated through time in a recurrent manner. The right image illustrates the individual components of the unified cell. This comprises of multi-image, single-image, residual and reconstruction components. The feature space outputs of the multi-image and single-image components are combined through element-wise subtraction and further processed, the results of which are added (element-wise) to the single image output to generate a super-resolved image. The network is fully convolutional and can take any input size. A given trained network can then up-sample with a fixed scale; i.e., 2x or 4x.


The proposed network integrates recurrent-based MISR and SISR, drawing benefits from both the methods to super-resolve thermal images. The single image network can be replaced by any state-of-the-art network but, currently, we have adopted the DBPN \cite{haris2018deep}. The MISR and Residual components include series of residual blocks. That is, convolutional layers with residual connections between block input and output. The reconstruction component is simply a convolutional layer that takes a combination of the outputs of the Residual component and SISR component to produce the super-resolved frame.

The complexity of our neural network architecture was analysed using 3 different metrics: The number of parameters, the number of multiply-accumulates (MACs) and the number of floating-point operations (FLOPs). The 4x version of the network had 5,151,645 parameters. For an 80$\times$80 input thermal image, this network performs 285.45 GMACs and 570.91 GFLOPs.

\subsection{Training}
\vspace{-0.1cm}
\begin{wrapfigure}{r}{0.55\textwidth}

\vspace{-0.5cm}
\includegraphics[width=0.55\textwidth]{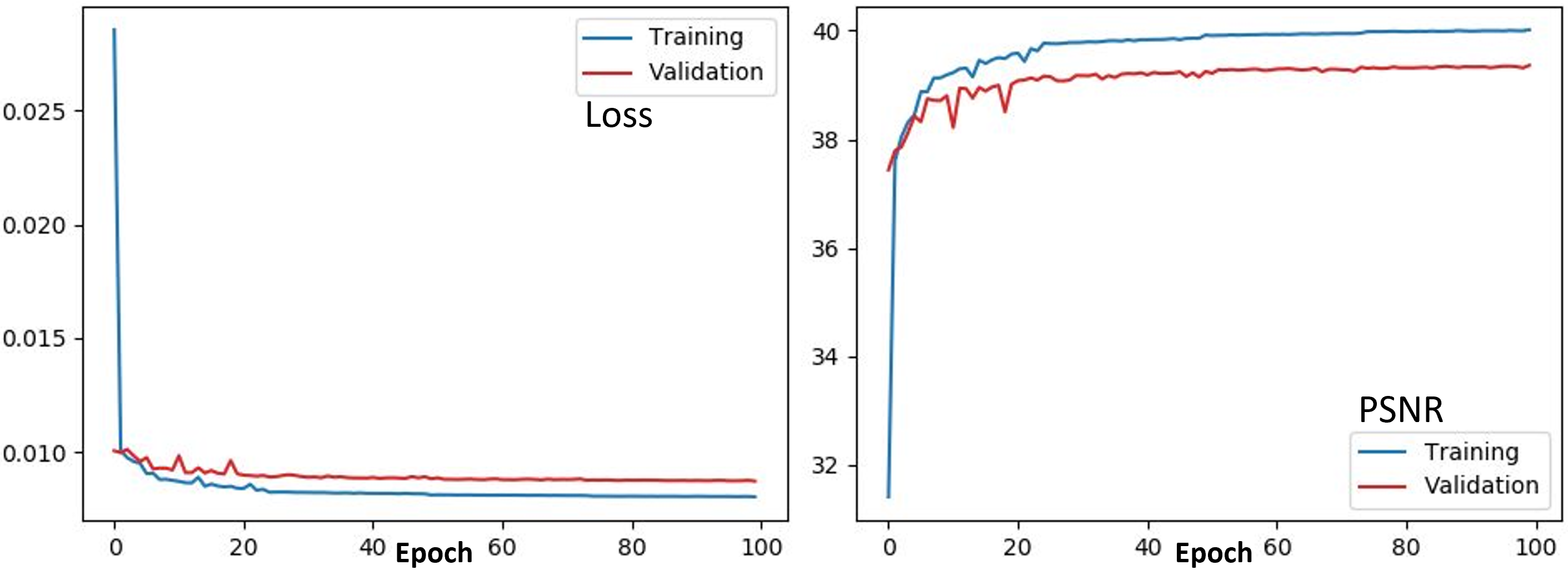}
\vspace{-0.7cm}
\caption{Progression of mean absolute error loss and PSNR during training.}
\label{fig:loss}
\vspace{-0.2cm}
\end{wrapfigure}

The network was trained using PyTorch,the deep learning framework. The sequence length of each sample passed to the network during training was randomised between 1 and 10 to allow the recurrent network to generalise to varying lengths of input. The same 128$\times$128 random crop was taken from each HR frame in the sequence and the LR frames were created by down-sampling by a factor of 4 and applying Gaussian blur and additive noise to simulate degradation due to the camera resolution. The ADAM optimizer was used with a learning rate of $10^{-4}$ reduced by a factor of 0.5 every 25 epochs for a total of 100 epochs. Mean absolute error was used as the loss function and a weight decay of $10^{-4}$ was applied. No form of pre-training was done. Figure \ref{fig:loss} shows the loss and PSNR progression on the training and validation data during training.



\section{Experimental Results}
\begin{wraptable}{r}{0.45\textwidth}
\centering
\vspace{-0.6cm}
\vspace{0.1cm}
\begin{tabular}{lcc}
\hline
        & PSNR            & SSIM          \\ \hline
SR      & 39.235 ± 2.639  & 0.901 ± 0.051 \\ 
Bicubic & 37.705 ± 1.952  & 0.879 ± 0.045 \\ \hline
\end{tabular}
\label{tab:results_table}
\caption{Performance on validation dataset.}
\vspace{-0.2cm}

\end{wraptable}
The performance of the proposed super-resolution network trained on our collected thermal dataset was evaluated using PSNR and SSIM as quantitative metrics. The mean values of these metrics on our validation dataset using our trained SR network and bicubic interpolation are shown in Table \ref{tab:results_table}.

\begin{figure}[htbp]
\centerline{\includegraphics[width=0.75\textwidth]{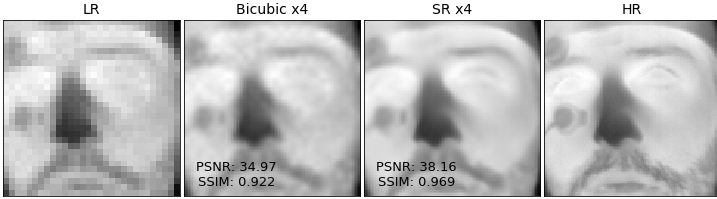}}
\vspace{-0.1cm}
\centerline{\includegraphics[width=0.75\textwidth]{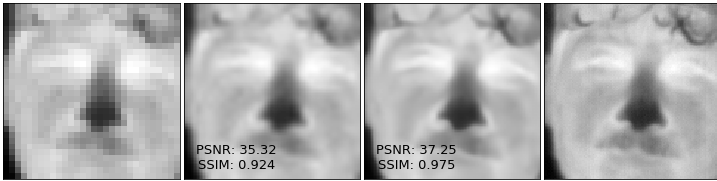}}
\vspace{-0.4cm}
\caption{Examples of SR network performance on sample images from our validation set.}
\label{fig:thermal_sr_example1}
\vspace{-1cm}
\end{figure}

The proposed SR network outperforms bicubic interpolation significantly in both of these metrics and also does so visually. Qualitative examples of our SR networks performance can be visualized in Figure \ref{fig:thermal_sr_example1}. 

\vspace{-0.3cm}

\section{Conclusions}
In this study, we propose a novel multi-image super-resolution architecture for in-cabin driver monitoring systems that exhibits exemplary performance on a locally acquired thermal facial dataset acquired from an uncooled thermal camera. The proposed network integrates recurrent-based MISR and SISR. The trained network (PSNR: 39.24, SSIM: 0.90) outperforms bicubic interpolation (PSNR: 37.71, SSIM: 0.88) both quantitatively and qualitatively. As part of future work, we plan to optimise the network to further accelerate the inference speed thus making it more efficient and computationally less expensive for real-time deployment on low-power edge computing devices (such as NVIDIA-Jetson). We also plan to retrain the network on a larger and more diverse dataset.






\bibliographystyle{apalike}
\vspace{-0.3cm}
\bibliography{references}

\end{document}